\documentclass[conference]{IEEEtran}
\IEEEoverridecommandlockouts
\usepackage[utf8]{inputenc}
\usepackage{graphicx}

\graphicspath{ {imagesConf/} } %%%This is the path for images, and we skip adding the long URL for images' path.
\makeatletter
\newcommand{\linebreakand}{%
  \end{@IEEEauthorhalign}
  \hfill\mbox{}\par
  \mbox{}\hfill\begin{@IEEEauthorhalign}
}
\makeatother

\usepackage{caption}
\usepackage{subcaption}
\usepackage{cite}
\usepackage{url}
\usepackage{amsmath,amssymb,amsfonts}
\usepackage{algorithmic}
\usepackage{graphicx}
\usepackage{textcomp}
\usepackage{xcolor}
\def\BibTeX{{\rm B\kern-.05em{\sc i\kern-.025em b}\kern-.08em
    T\kern-.1667em\lower.7ex\hbox{E}\kern-.125emX}}
    
\usepackage{todonotes}

\usepackage{multirow}
\usepackage{color, colortbl}
\usepackage{booktabs, caption, makecell}

\usepackage{threeparttable}
\usepackage{float}

\usepackage{balance}
\usepackage{lastpage}

\definecolor{my-grey}{RGB}{238,238,238}
\newcommand*{\belowrulesepcolor}[1]{%
  \noalign{%
    \kern-\belowrulesep
    \begingroup
      \color{#1}%
      \hrule height\belowrulesep
    \endgroup
  }%
}
\newcommand*{\aboverulesepcolor}[1]{%
  \noalign{%
    \begingroup
      \color{#1}%
      \hrule height\aboverulesep
    \endgroup
    \kern-\aboverulesep
  }%
}
\newif\ifblackandwhite
\usepackage[hmargin=2cm,vmargin=2.5cm]{geometry}
\usepackage{etoolbox}
\usepackage{longtable}%
\AtBeginEnvironment{longtable}{%
}
\ifblackandwhite
  
\else

\fi
\usepackage{adjustbox}
\makeatletter
 \let\old@ps@headings\ps@headings
 \let\old@ps@IEEEtitlepagestyle\ps@IEEEtitlepagestyle
 \def\confheader#1{%
 \def\ps@IEEEtitlepagestyle{%
 \old@ps@IEEEtitlepagestyle%
 \def\@oddhead{\strut\hfill#1\hfill\strut}%
 \def\@evenhead{\strut\hfill#1\hfill\strut}%
 }%
 \ps@headings%
 }
 \makeatother

\begin{document}

\title{Firearm Detection and Segmentation Using an Ensemble of Semantic Neural Networks}
% Firearm Detection via Semantic Segmentation

%{\footnotesize \textsuperscript{*}Note: Sub-titles are not captured in Xplore and should not be used}
%\thanks{Identify applicable funding agency here. If none, delete this.}%}

\author{\IEEEauthorblockN{Alexander Egiazarov}
\IEEEauthorblockA{\textit{Digital Security Group} \\
\textit{University of Oslo}\\
Oslo, Norway  \\
alexaeg@ifi.uio.no}
\and
\IEEEauthorblockN{Vasileios Mavroeidis}
\IEEEauthorblockA{\textit{Digital Security Group} \\
\textit{University of Oslo}\\
Oslo, Norway \\
vasileim@ifi.uio.no}
\linebreakand
\IEEEauthorblockN{Fabio Massimo Zennaro}
\IEEEauthorblockA{\textit{Digital Security Group} \\
\textit{University of Oslo}\\
Oslo, Norway \\
fabiomz@ifi.uio.no}
\and
\IEEEauthorblockN{Kamer Vishi}
\IEEEauthorblockA{\textit{Digital Security Group} \\
\textit{University of Oslo}\\
Oslo, Norway \\
kamerv@ifi.uio.no}
\linebreakand
}

\maketitle

\begin{abstract}
In recent years we have seen an upsurge in terror attacks around the world. Such attacks usually happen in public places with large crowds to cause the most damage possible and get the most attention. Even though surveillance cameras are assumed to be a powerful tool, their effect in preventing crime is far from clear due to either limitation in the ability of humans to vigilantly monitor video surveillance or for the simple reason that they are operating passively. In this paper, we present a weapon detection system based on an ensemble of semantic Convolutional Neural Networks that decomposes the problem of detecting and locating a weapon into a set of smaller problems concerned with the individual component parts of a weapon. This approach has computational and practical advantages: a set of simpler neural networks dedicated to specific tasks requires less computational resources and can be trained in parallel; the overall output of the system given by the aggregation of the outputs of individual networks can be tuned by a user to trade-off false positives and false negatives; finally, according to ensemble theory, the output of the overall system will be robust and reliable even in the presence of weak individual models. We evaluated our system running simulations aimed at assessing the accuracy of individual networks and the whole system. The results on synthetic data and real-world data are promising, and they suggest that our approach may have advantages compared to the monolithic approach based on a single deep convolutional neural network.

\end{abstract}


\section{Introduction}
%(Relevance of the problem; our contribution)
%(Couple of paragraphs, around .5 pages)

According to the report \textit{"Global Study on Homicide 2019"}, published by the United Nations Office on Drugs and Crime (UNODC), more than 464,000 people across the world were killed in homicides in 2017, accounting for more than five times of the number of people killed in armed conflict.  Moreover, excluding homicides of unknown mechanism, more than half  (roughly  54\%) of homicides worldwide in 2017 were perpetrated with a firearm, and slightly more than a  fifth  (22\%) were perpetrated with a sharp object \cite{UNODC2019}. The use of technology in law enforcement and crime prevention is developing rapidly. For example, video surveillance technology has been proved useful in aiding investigators either to identify new suspects related to a crime or to bolster evidence against current ones. In some cases, video footage can be a decisive piece of evidence in the course of a trial. Current video surveillance systems (also known as CCTV) are generally operating passively \cite{Darker2017_CCTV}, which indicates that related authorities usually become aware of a crime event postliminary. For example, CCTV operators or police officers review a video footage only after a crime occurred. In addition, in cases where video surveillance technology is used for real time monitoring as a way of deterring potential crimes, an agent would have to visually detect the presence of weapons in the monitored scenes and make decisions in a very short time. Any delay or miscalculation in the decision making process of an agent could be detrimental and result even in loss of human lives. One of the most effective solutions to this problem is to enhance surveillance cameras with Artificial Intelligence (AI), where an accurate proactive crime-fighting tool can be used to detect and consequently prevent weapon crimes. 

The contribution of this research is twofold. First, we apply the concept of semantic decomposition to the problem of detecting a weapon; that is, we reduce the holistic problem of detecting a whole weapon to the subset of smaller, semantically-related problems of identifying the different component parts of a weapon. 
%modeling where the classification of a concept is not handled holistically (atomically) but modularly providing an approach for solving high-level problems through a set of similar smaller ones. %i can say here why is this useful.
Second, we apply and evaluate this approach on a weapon detection system that we developed using deep Convolutional Neural Networks (CNNs) for detecting individual parts of a firearm in images and videos.

The proposed method can be used in several applications, such as for real-time detection of weapons in places where video surveillance is available or needed (e.g., airports, stadiums, other public places), for control purposes  (e.g., in the case of videos or images uploaded to social media containing graphics with weapons), and for complementing other current security technologies (e.g.,  metal detectors that can detect weapons concealed on an individuals body).

%%%%-->PAPER STRUCTURE SECTION
The paper is structured as follows. Section \ref{sec:Background} presents background information and related work. Section \ref{sec:ProblemStatement} formalises the goal of our research and our approach to solving the problem of weapon detection via machine learning. Section \ref{sec:DataGeneration} analyses the dataset generation protocol for our study, while Section \ref{sec:Model} provides details about the model we implemented. Section \ref{sec:ExperimentalValidation} presents different simulations conducted for validating our model and assessing its performance. Section \ref{sec:EthicalConsiderations} discusses some ethical implications related to our work. Finally, Section \ref{sec:Conclusions} concludes our work and outlines future developments. 

\section{Background \label{sec:Background}}
%(Short info of the background, relevant work and state of the art in the areas concerned)
%(Organization of the paper)
%(around 1 pages, .5 per subsection)

\subsection{Security}
%(Problem of detecting weapons)

No region in the world is exempt from the dramatic consequences of firearms violence. The problems associated with firearms violence covers the whole spectrum of human security: ranging from high levels of individual physical insecurity (domestic and street violence, gang and criminal violence) with severe economic and social consequences for the society at large, to large scale armed conflicts in which these arms enable widespread violence and account for the majority of deaths.

\subsection{Machine learning}
Machine learning provides a set of methods and techniques for inferring patterns and relations among data. 
\emph{Supervised learning} is the task of learning a functional relationship between a set of data and a semantic class. Given a set of samples $\mathbf{x}$ as vectors, where each of the vectors is associated with a label $y$, the aim of a machine learning algorithm is to discover a function $f: \mathbf{x} \mapsto y$ that associates every sample to its label \cite{bishop2006pattern}.

A popular family of algorithms for supervised learning is \emph{neural networks}, which constitute universal approximators that can be effectively trained by backpropagation. \emph{Convolutional Neural Networks}, in particular, are a sub-set of neural networks designed to process two-dimensional inputs and be invariant to translations; these networks can be used to process images and to carry out tasks such as image detection or image segmentation.

The availability of large amounts of data and computational power has recently fostered the development of \emph{deep learning} algorithms, that is, the use of layered neural networks able to learn complex functions \cite{lecun2015deep}. In particular, the success of deep CNNs on several image recognition tasks \cite{krizhevsky2012imagenet} has made these models a common choice for solving problems of image recognition, detection and segmentation. Starting from the simple pixel representation of an image, deep CNNs automatically discover robust and increasingly higher-level features that can be used to solve challenging problems in image recognition and segmentation \cite{GUO201627,Krizhevsky2012}.

\subsection{Related work}
The problem of weapon detection in videos or images using machine learning is related to two broad research areas. The first research area focuses on addressing weapon detection using classical computer-vision or pattern-matching methods, whereas the second research area focuses on improving the performance of object detection using machine learning techniques.

\begin{itemize}
    \item \textbf{Weapon detection:} The majority of methods used for weapon detection are based on imaging techniques, such as infrared imaging \cite{Xue2002} and millimeter wave imaging \cite{Sheen2001}. The most representative application in this context is luggage control in airports.
    
    %%%Maybe too detailed related work. THESE PARAGRAPHS ARE HERE IF WE WANT TO MAKE IT DETAILED!
    
    %%D. M. Sheen et al. \cite{Sheen2001} proposed a method of CWD for airports and other secure location based on three-dimensional millimeter (mm) wave imaging technique. 
    %%Z. Xue et al. \cite{Xue2002} proposed a method of CWD based on fusion of infrared (IR) imaging and color visual image. The authors use the fusion of infrared image and visual image method to maintain the natural colour of the original image. 
    %%R. Blum et al. \cite{Blum} developed a method of CWD based on fusion of IR or mm-wave image and visual image using a multi-resolution mosaic technique where a concealed weapon is first detected by fuzzy k-means clustering method from IR or mm-wave image. 
    %%E. M. Upadhyay et al. \cite{Upadhyay} proposed a method of CWD using image fusion. The authors used a homomorphic filter, entropy of blocks and blending approach of fusion to generate a multi exposure-multi modal image from a set of visual and IR image with multiple exposures.

    \item \textbf{Object segmentation models:} Object segmentation consists of recognizing an object and finding its location in the input image. The existing methods address the segmentation problem by reformulating it into a classification problem; first a classifier is trained, and then, during the detection process, filters are run on a number of areas of the input image using either the \textit{sliding window} approach \cite{Dalal2005_top,Felzenszwalb2010,felzenszwalb2005pictorial} or the \textit{region proposals} approach \cite{Hosang2016,Girshick2014,uijlings2013selective,NIPS2015_5638}. 
\end{itemize}

Standard machine learning approaches based on deep convolutional neural networks have also been used to try to solve the problem of detection and segmentation at the same time \cite{gelana2019firearm,olmos2018automatic}. However, to the best of our knowledge this is the first work focusing on firearm detection and segmentation using an ensemble of small semantic neural networks.

\section{Problem Statement \label{sec:ProblemStatement}}
%(Problems we face: (i) absence of data; (ii) limits of standard approach...)
%(Clear, exact definition of what we are trying to do)
%(.5 pages)
In this paper, we tackle the problem of detecting the presence of weapons within an image. We cast this problem as image detection/segmentation supervised learning problem: we want to learn a function that given an image as input returns the presence and the location of the weapon. Following the state of the art, we decided to use convolutional neural networks to learn such a function.

Weapon detection presents some peculiar and significant challenges. First of all, one of the main drivers of deep learning, that is the availability of data, is weak in this specific application domain. Differently from other standard image recognition problems that can exploit the availability of large standardized public datasets, we can not rely on similar resources. For this reason, we decided to collect and develop our own dataset (Section \ref{sec:DataGeneration}).

We also decided to consider a possible limitation on the other drive of deep learning, that is the availability of significant computational resources. To face this constraint, we decided to exploit common prior semantic knowledge about weapons: since any modern firearm is composed of different identifiable parts, we decided to decompose the overall problem of detecting a single weapon in a set of smaller problems concerned with the detection of the individual component parts of the weapon itself. We thus instantiated a set of small convolutional neural networks, each one solving the simpler problem of detecting a specific component part of a weapon, and aggregated the results in a robust output (Section \ref{sec:Model}). The individual networks are faster to train, requiring less computational resources and can be easily parallelized.

Training a set of smaller and potentially weaker learners finds justification in ensemble theory. It is well-known that simple weak learners (i.e., modules whose accuracy is just above a random chance threshold) may be effectively aggregated in ensemble modules whose accuracy is stronger than the individual learners \cite{rokach2010pattern}. This principle informs our hypothesis that aggregating potentially weak learners in the form of small dedicated networks detecting specific parts of a weapon may not only guarantee a computational advantage but also produce an ensemble system that is more robust and reliable than a single deep network.
%Second, since we are dealing with a sensitive security issue, a strict control over the mistakes of the network is crucial; we address this concern by enriching our training process with a human-in-the-loop (Section \ref{sec:Human-in-the-Loop}). 

\section{Dataset Generation \label{sec:DataGeneration}}
%(Technical explanation of how the dataset was created. Justify your choices. Link to your dataset if you make it available.)
%(If possible, within .5 pages)

\begin{figure}[h]
\centering
\caption{Main components of AR-15 rifle. \label{fig:components}}
\includegraphics[width=\columnwidth]{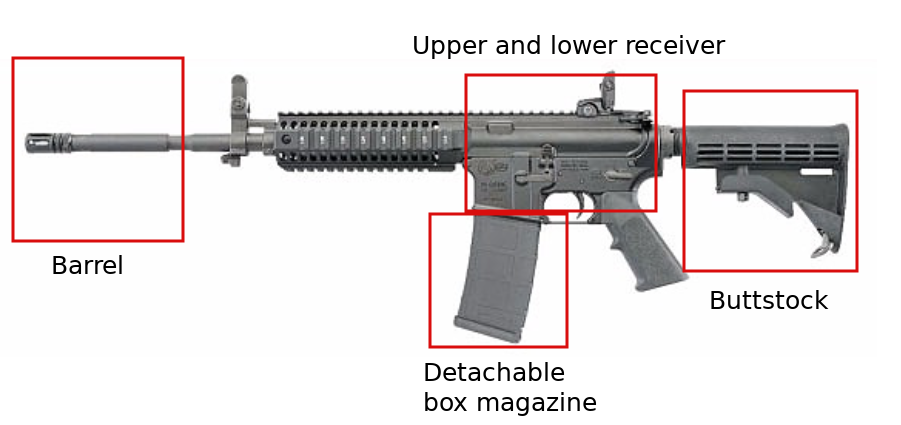}
\end{figure}

Due to the lack of comprehensive image libraries for firearms, we created a custom dataset. We decided to focus on a single specific type of rifle, AR-15 style rifle. The choice was made considering the popularity of this firearm in the Western part of the world, accessibility of the source images and visual diversity of different models with core features being preserved. It is worth noting that this type of firearm in addition to other rifles was used in terror attacks (mass shooting), such as in New Zealand in 2018 \cite{NEWZEALAND2018}, in Las Vegas, USA in 2017 \cite{LASVEGAS2017} and in Orlando, USA in 2016 \cite{ORLANDO2016}. We identified four essential component parts: barrel, magazine, butt-stock and upper-lower receiver assembly (see Figure \ref{fig:components}).

\begin{figure}[h]
\centering
\caption{An illustrative example of the stock and a negative sample. \label{fig:negative_instance}}
    \begin{subfigure}{0.48\columnwidth}
        \includegraphics[width=\linewidth]{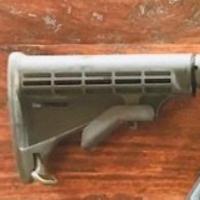}
    \end{subfigure}%
      \hfill %%
    \begin{subfigure}{0.48\columnwidth}
        \includegraphics[width=\linewidth]{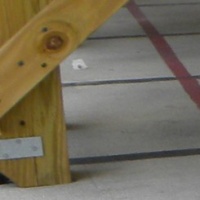}
    \end{subfigure}%
\end{figure}

To generate positive instances for our dataset we queried the Google images service and retrieved 4500 instances of pictures depicting AR-15 rifles. We first processed these images manually, generating and classifying 2500 samples for each component part. Following the standards \cite{bishop2006pattern}, we partitioned samples into training (2000 samples), validation (400 samples) and test (100 samples) sets; we then augmented our training dataset through synthetic manipulations (e.g., rotation, shearing, and scaling) to increase the total number of training samples per part to 8000.
To generate negative instances, we reused the initial source set of 4500 images and extracted random samples that explicitly do not contain any firearm parts, such as background details, humans, clothing and miscellaneous items present in the source images (see for an example Figure \ref{fig:negative_instance}).

In total, we have five datasets, one for each component plus one composed of exclusively negative examples; all the images add up to 42500 samples. Each picture is 200$\times$200 pixels.

\section{Semantic Segmentation Model \label{sec:Model}} 
%(Details of the model(s) implemented. Within the limits, this should be detailed enough to make your work reprodubicle; report architectures and hyperparameters whenever possible. Link to the source code if you make it available)

One of the main contributions of this work is the semantic decomposition of the detection problem. Instead of directly solving the challenge of detecting a weapon, we developed a system that relies on the identification of its parts. Breaking down the problem of identifying a weapon into the problem of identifying its components simplifies the problem at hand: it provides a divide-and-conquer approach to solve a high-level problem through the solution of a set of smaller similar problems. Moreover, in this case, it also offers a simple way to deal with situations where firearms are partially concealed, lack certain distinct features or are modified.

\begin{figure}[h]
\caption{Neural network architecture. \label{fig:NN-architecture}}
\includegraphics[width=\linewidth]{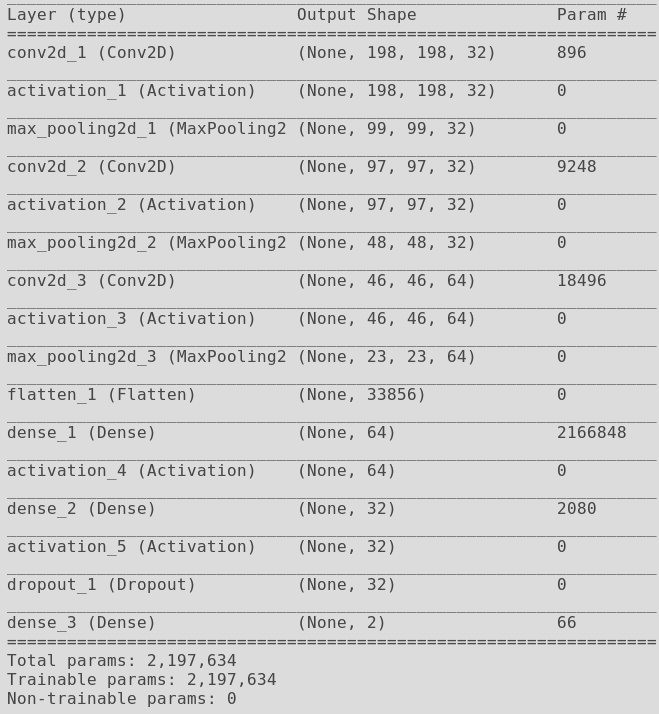}
\end{figure}

We implement a modular array of convolutional neural networks\footnote{All models are implemented using Keras (\url{https://keras.io/}) and Tensorflow (\url{https://www.tensorflow.org/})}. All the networks have the same architecture: three convolutional layers for feature extraction and three dense layers for binary classification (see Figure \ref{fig:NN-architecture} for reference).
In the convolutional section, we use layers containing 32 or 64 filters with default stride of 1x1, with ReLU activation functions, and 2x2 max-pooling.
In the dense section, we use fully connected layers. We use a ReLU activation function, except for the last layer where we rely on the softmax function to compute the output. Moreover, in the second-to-last layer, we use dropout \cite{Srivastava2013a} with a probability $p=0.5$ for regularization. 
The relatively small architecture allows for faster parallel tuning and training.

At runtime, in order to process an image of arbitrary size, a simple iterative algorithm based on a sliding window is used to generate cropped images of the required input size of $200 \times 200$ pixels.

We take the output of each network to carry equal weight towards the final detection decision, each one contributing 25 percent certainty about the presence of a firearm.
However, the mechanism to aggregate each output in a final decision can be easily changed by the developer of the system by tweaking the contribution weight of each component of the weapon and the number of components that should be detected before raising an alert: in sensitive and critical situations, the system can be configured to raise a firearm detection alert whenever a single component part is identified; in less critical scenarios, the sensitivity (and, relatedly, the number of false positive detections) may be lowered by requiring the identification of multiple component parts before raising an alert. Notice, also, that a developer may easily decide to instantiate and aggregate different custom-designed networks; the modular design of our system allows for straightforward plugin and removal of additional networks, whose output is aggregated in the final decision.

Finally, to assess image segmentation, the outputs of the individual networks are used to produce a heatmap and a bounding box marking the region of the image where our model returned a high positive activation, thus pinpointing the location of the detected weapon in the image.

\section{Experimental Validation \label{sec:ExperimentalValidation}}
%(Core of the article where to report experiments and results)
%(Between 3-5 pages with the most important results)
%(Each subsection Simulation N should report (at least) protocol, results and discussion. The division in subsubsection may be discarded, but it helps structuring your paper. One of them should be the overall evaluation of the modules. Then choose the most interesting/representative experiments: sensitivity to colours, texture recognition, occlusion, proof-of-concepts...)

\subsection{Simulation 1: Quantitative assessment of the individual networks}
The first simulation assesses the performance of each component network on the test data.

\paragraph{Protocol}
After training and validating our networks, we tested each of them on the corresponding test data set composed of 100 positive and 100 negative instances. The aim of this simulation is to evaluate the accuracy achieved on the test set and to identify and understand the potential weaknesses of the system by examining the cases on which the system returned a wrong output.
The output layer of the network has two nodes, one activated when the network believes a weapon is present, and one activated when the network believes a weapon is absent. Since the last layer is a softmax, the total activation of the layer is one. The threshold for weapon detection is set at 0.5. If the output of the node for the presence of a weapon is greater than 0.5, then we have a positive output; otherwise, we have a negative output.

\paragraph{Results}
Table \ref{tab:perfromance1} reports the absolute detection values on positive and negative samples for each component network along with the relative accuracy, precision and recall.
%%%%%%%%%TABLE 1%%%%%%%%%%
%\begin{table}[!htbp]
%\centering
%\caption{Detection performance for each network of each component. \label{tab:perfromance1}} 
%\begin{adjustbox}{width=1\columnwidth}
%\begin{threeparttable}
%\begin{tabular}{lllccc}
%\toprule
%\belowrulesepcolor{my-grey}\rowcolor{my-grey}
%		\textbf{Part name} & \textbf{Pos. set} & \textbf{Neg. set} & \textbf{Accuracy} & \textbf{Precision} & \textbf{Recall} \\ 
%		\aboverulesepcolor{my-grey}
%		\midrule
%		Stocks        & \begin{tabular}[c]{@{}l@{}}TP: 463\\ FP: 57\end{tabular} & \begin{tabular}[c]{@{}l@{}}TN: 413\\ FN: 87\end{tabular}  & 0.876 & 0.89 & 0.842 \\ \hline
%		Magazines     & \begin{tabular}[c]{@{}l@{}}TP: 387\\ FP: 113\end{tabular} & \begin{tabular}[c]{@{}l@{}}TN: 417\\ FN: 83\end{tabular} & 0.804 & 0.74 & 0.823 \\ \hline
%		Barrels       & \begin{tabular}[c]{@{}l@{}}TP: 485\\ FP: 15\end{tabular} & \begin{tabular}[c]{@{}l@{}}TN: 390\\ FN: 110\end{tabular} & 0.875 & 0.97 & 0.815 \\ \hline
%		Receivers     & \begin{tabular}[c]{@{}l@{}}TP: 418\\ FP: 82\end{tabular}  & \begin{tabular}[c]{@{}l@{}}TN: 489\\ FN: 11\end{tabular} & 0.907 & 0.836 & 0.974\\ \hline
%	\end{tabular}
%	\begin{tablenotes}\footnotesize
%\item[*] \textbf{Note:} \emph{TP} stands for true positives, \emph{FP} for false positives, \emph{TN} for true negatives, and \emph{FN} for false negatives.
%\end{tablenotes}
%	\end{threeparttable}
%	\end{adjustbox}
%\end{table}
%%%%%%%%%%%%%%%END OF TABLE 1 %%%%%%%%%%%%%%

%%%%%%%%%TABLE 2%%%%%%%%%%
\begin{table}[!htbp]
\centering
\caption{Detection performance for each network of each component. \label{tab:perfromance1}} 
\begin{adjustbox}{width=1\columnwidth}
\begin{threeparttable}
\begin{tabular}{lllccc}
\toprule
\belowrulesepcolor{my-grey}\rowcolor{my-grey}
		\textbf{Part name} & \textbf{Pos. set} & \textbf{Neg. set} & \textbf{Accuracy} & \textbf{Precision} & \textbf{Recall} \\ 
		\aboverulesepcolor{my-grey}
		\midrule
		Stocks        & \begin{tabular}[c]{@{}l@{}}TP: 97\\ FP: 3\end{tabular} & \begin{tabular}[c]{@{}l@{}}TN: 78\\ FN: 22\end{tabular}  & 0.875 & 0.97 & 0.815 \\ \hline
		Magazines     & \begin{tabular}[c]{@{}l@{}}TP: 81\\ FP: 19\end{tabular} & \begin{tabular}[c]{@{}l@{}}TN: 84\\ FN: 16\end{tabular} & 0.825 & 0.81 & 0.835 \\ \hline
		Barrels       & \begin{tabular}[c]{@{}l@{}}TP: 99\\ FP: 1\end{tabular} & \begin{tabular}[c]{@{}l@{}}TN: 54\\ FN: 46\end{tabular} & 0.765 & 0.99 & 0.683 \\ \hline
		Receivers     & \begin{tabular}[c]{@{}l@{}}TP: 82\\ FP: 18\end{tabular}  & \begin{tabular}[c]{@{}l@{}}TN: 95\\ FN: 5\end{tabular} & 0.885 & 0.82 & 0.943\\ \hline
	\end{tabular}
	\begin{tablenotes}\footnotesize
\item[*] \textbf{Note:} \emph{TP} stands for true positives, \emph{FP} for false positives, \emph{TN} for true negatives, and \emph{FN} for false negatives.
\end{tablenotes}
	\end{threeparttable}
	\end{adjustbox}
\end{table}
The network responsible for the detection of receivers achieved 95\% accuracy when tested against negative samples, but performed slightly worse in detecting the actual receivers achieving 82\% accuracy. The network responsible for the detection of magazines achieved 84\% accuracy against the negative samples and 81\% accuracy in detecting magazines (positive samples). The network trained on barrels correctly classified 54\% of the negative samples, and performed remarkably well in detecting barrels with an accuracy of 99\%. The results of the network responsible for the detection of the stocks gave an 78\% accuracy in classifying correctly negative samples and 97\% accuracy in detecting stocks.

\paragraph{Discussion}
Overall, the performance of the individual networks is satisfying, with an accuracy ranging between 75 and 90 percent. Notice that the strength of our system relies not only on the performance of the individual networks but also on the ensembling of the results of all networks. For instance, assume that the system is set up to raise a positive detection alert if at least two of the networks output a positive signal; assume also that the average accuracy on positive samples is 0.8 (as in our simulations); then, the overall probability of the system of missing a positive sample is given by the probability of three individual networks of making a mistake in recognizing a positive sample, that is, $(1-0.8)^3 = 0.008$; the overall system may then reach an accuracy up to $0.992$ in detecting positive samples\footnote{Notice, though, that this value of accuracy is an upper bound based on simplifying assumptions, such as the assumption of independence of the outputs of each network and the assumption that each part is always detectable from each image. Such assumptions are likely not to hold in real-world scenarios.}. This result is consistent with the ensemble theory discussed in Section \ref{sec:ProblemStatement}.

\subsection{Simulation 2: Evaluation of semantic segmentation}
In the second simulation, we further assess the behaviour of the individual neural networks assembled into a coherent system and provide a proof of concept for the principle of semantically decompositon of the problem of weapon detection.

\begin{figure}[h]
\centering
\caption{Source image, background and a sample combined test image. \label{fig:second-simul}}
    \begin{subfigure}{0.48\columnwidth}
        \includegraphics[width=\linewidth]{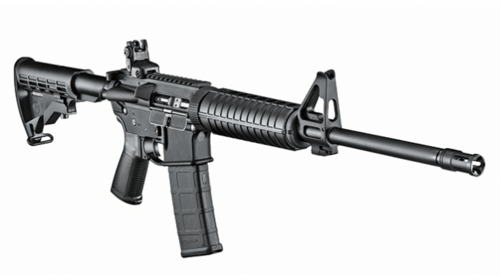}
    \end{subfigure}
    \hfill
    \begin{subfigure}{0.48\columnwidth}
        \includegraphics[width=\linewidth]{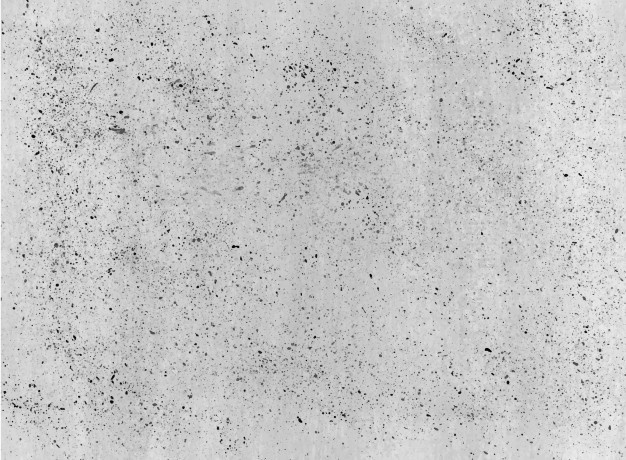}
    \end{subfigure}
    \begin{subfigure}{0.48\columnwidth}
        \includegraphics[width=\linewidth]{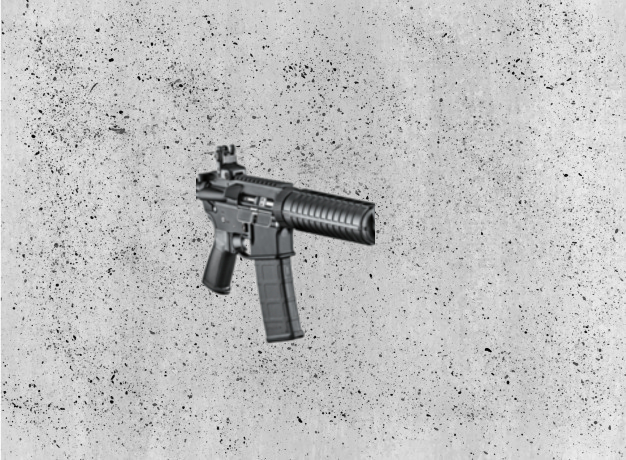}
    \end{subfigure}
\end{figure}

\paragraph{Protocol}
We tested the reliability of the system by feeding images (from the datasets) of AR-15 style rifle on a noisy neutral background (e.g., an image of a concrete surface). We selected a high-quality image of the weapon and we artificially produced different sets of transformed images, such as \emph{rigid transformations of the weapon} (change of orientation, scale and location) and \emph{ablative transformations of the weapon} (removal and occlusion of component parts). We ended up working with both images of single component parts and images of the whole rifle. In total, we generated 14 random images that allow us to evaluate both the performance of individual networks (as we did in the previous simulation) and the performance of the whole system. For a sample of a used image, the background, and a sample ablative transformation, see Figure \ref{fig:second-simul}.

The setup of this simulation allows us to evaluate the performance of the network independently of possible confounding elements in the background. In other words, it allows us to assess whether the networks are able to identify their target component both in an ideal setting and under several challenging transformations of the weapon.

We evaluate the results of our simulations by assessing the detection and segmentation performance of our system. The detection is evaluated by checking whether a specific network (in case of an image of a component) or the ensemble system (in case of an image of the whole weapon) produce a positive output. The segmentation is evaluated by computing a bounding box and checking it against a manually defined ground-truth bounding box. A segmentation is deemed successful if the area of the bounding box covers the firearm or all visible parts. If the bounding box is clearly outside of the boundaries of the firearm or if it is covering a sizeable non-firearm area, then the segmentation is deemed unsuccessful.

\begin{figure}[h]
\centering
\caption{Results of the segmentation on sanitized images. The green bounding box is the manually labeled area, whereas the blue bounding box is the area detected by the networks. \label{fig:simul2-part}}
    \begin{subfigure}{0.48\columnwidth}
        \includegraphics[width=\linewidth]{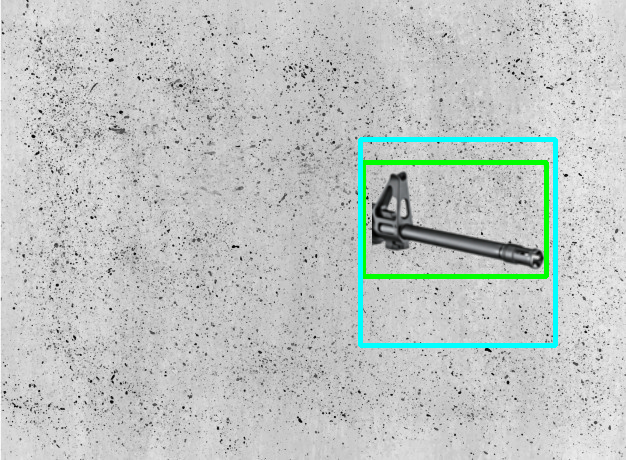}
    \end{subfigure}
    \hfill
    \begin{subfigure}{0.48\columnwidth}
        \includegraphics[width=\linewidth]{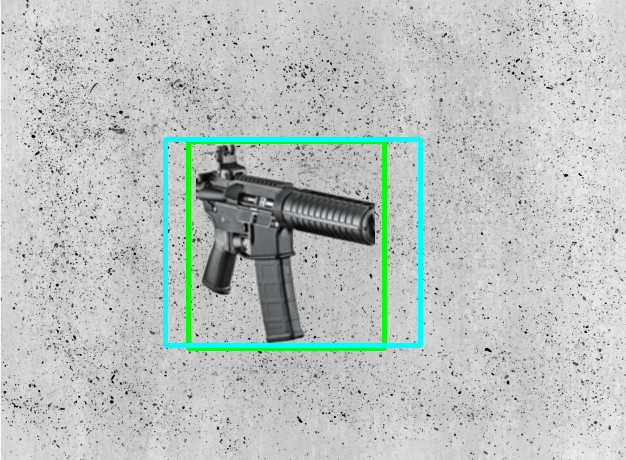}
    \end{subfigure}
    \hfill
    \begin{subfigure}{0.48\columnwidth}
        \includegraphics[width=\linewidth]{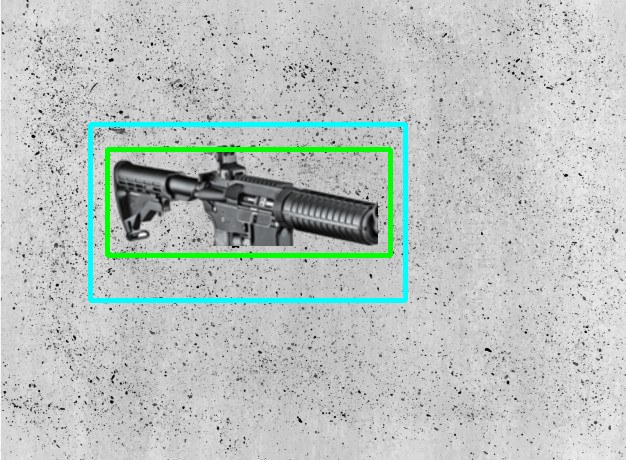}
    \end{subfigure}
    \hfill
    \begin{subfigure}{0.48\columnwidth}
        \includegraphics[width=\linewidth]{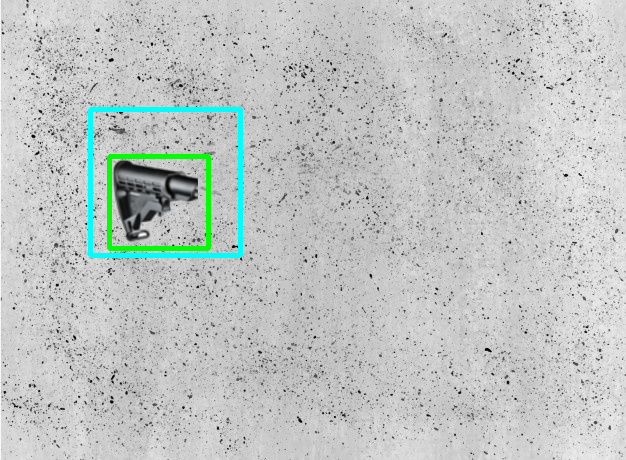}
    \end{subfigure}
\end{figure}

\begin{figure}[h]
\caption{Bounding box computed by aggregation of the outputs from individual networks. The green bounding box is the manually labeled area, whereas the blue bounding box is the area detected by the networks. \label{fig:simul2-whole}}
\includegraphics[width=\linewidth]{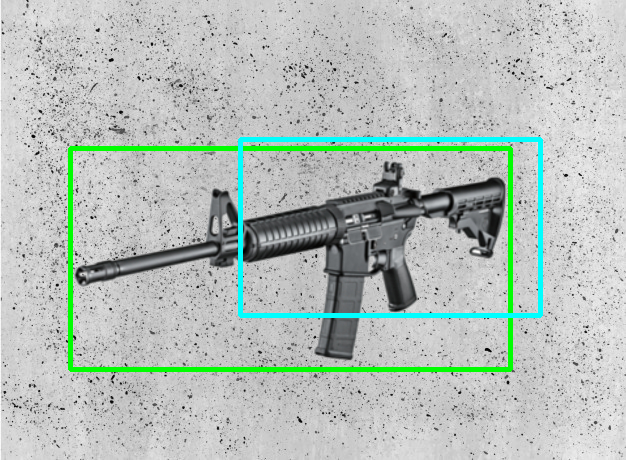}
\end{figure}

\paragraph{Results}
Testing 14 images, the system correctly detected the presence of a weapon, without generating any false negatives. The segmentation was also satisfactory, with the system successfully placing a bounding box over the firearm despite different orientations, scaling and part combinations. Figure \ref{fig:simul2-part} shows different sample bounding boxes computed by the system on components that are part of the rifle. Similarly, Figure \ref{fig:simul2-whole} shows the overall bounding box computed on an image of a rifle by aggregating individual bounding boxes. Notice that the end parts, such as the barrel, while correctly identified by the individual network, were sometimes not covered in the aggregated bounding box. The reason for that is that the aggregation algorithm weighted more overlapping areas such as the central parts of the weapon.

\paragraph{Discussion}
%(Interpretation, meaning and consequences of the results)
The results of this simulation are very encouraging: even though these tests were conducted on sanitized images in which potential distractors were removed, they demonstrate that the array of networks is robust enough in detecting parts of a weapon under different transformations.
Removing certain parts of the rifle proves the main strength of our solution based on a set of semantic networks: even when some parts of a firearm object are obscured or concealed, our system can still detect the presence of the firearm as long as at least some of the predefined parts are present. This makes our system suitable for deployment in environments where detecting concealed firearms in a rapid manner can mitigate potentially harmful events.

\subsection{Simulation 3: Evaluation on realistic video frames}
In this simulation we assess the accuracy of our system by processing images obtained from a custom video stream rather than images obtained from the Internet.

\paragraph{Protocol}
%(Specific setting of this simulation. Measures you use to evaluate it. Again, ideally it is reproducible.)
We processed a video stream containing a single standing person holding a AR-15 rifle. The subject is holding the rifle in three possible positions (low, medium and high). For each position, we extract 16 images showing the person holding the rifle in different degrees of rotation. The clothing of the subject and the background in the images are neutral to ensure that the rifle does not explicitly blend with the surroundings or the person. 
In total, we processed 48 sample images in which the weapon is presented in different rotation angles, and also situations in which certain parts of the rifle are obscured by the subject's clothing or the way the weapon is held.

%This setup allows us to further study the ability of the developed detection algorithm for locating and correctly classifying AR-15 type rifles, using realistic images as the ones that would be produced and processed coming by a video surveillance feed.

Similarly to previous simulations, we evaluate our results by computing the detection accuracy and presenting the bounding boxes of the identified objects. For each testing image, we provided a ground-truth bounding box by manually identifying the area in a frame where the rifle or parts of the rifle were visible. We quantitatively computed the overlap between the true bounding box and the predicted bounding box by evaluating the intersection between the true bounding box over the union of the bounding boxes computed by each one of our component networks. 

\paragraph{Results}
Table \ref{tab:perfromance2} reports the number of correct detections of the network and the number of instances where the computed bounding box overlapped at least 50 per cent the true bounding box. Images where the subject was holding the rifle at a low position, were successfully detected 13 times out of 16, even when the overlapping of bounding boxes was low (4 out of 16 with 50\% or more overlap). Images where the subject was holding the rifle at a medium position, performed slightly worse, with our system correctly classifying 10 out of 16 samples, whereas there was an improvement in accuracy of the segmentation boxes with 6 out of 16 boxes overlapping equally or more than 50\%. Finally, images where the subject was holding a rifle at a higher position our system achieved higher accuracy by correctly classifying 15 out of 16 images. In addition, 9 out of 16 images had their computed bounding boxes overlapping for 50\% or more with the true bounding boxes.
%%%%%%%%%TABLE 2%%%%%%%%%%%%
\begin{table}[h]
\centering
\caption{Detection and segmentation performance on the simulated video data \label{tab:perfromance2}}
\begin{adjustbox}{width=0.76\columnwidth}
\begin{tabular}{lcc}
\toprule
\belowrulesepcolor{my-grey}\rowcolor{my-grey}\textbf{Position} & \textbf{Detection} & \textbf{Overlap over 50\%} \\
\aboverulesepcolor{my-grey}
\midrule
Low               & 13 / 16                  & 4                          \\ \hline
Medium            & 10 / 16                 & 6                          \\ \hline
High              & 15 / 16                 & 9                          \\ 
\bottomrule
\end{tabular}
\end{adjustbox}
\end{table}
%%%%%%%%%%%%%%%%%%%%%%%%%%%%%%

\begin{figure}[h]
\caption{Result of the simulation testing with rifle being held in ``high" position. The green bounding box is the manually labeled area, whereas the blue bounding box is the area detected by the networks. \label{fig:video1}}
\includegraphics[width=\linewidth]{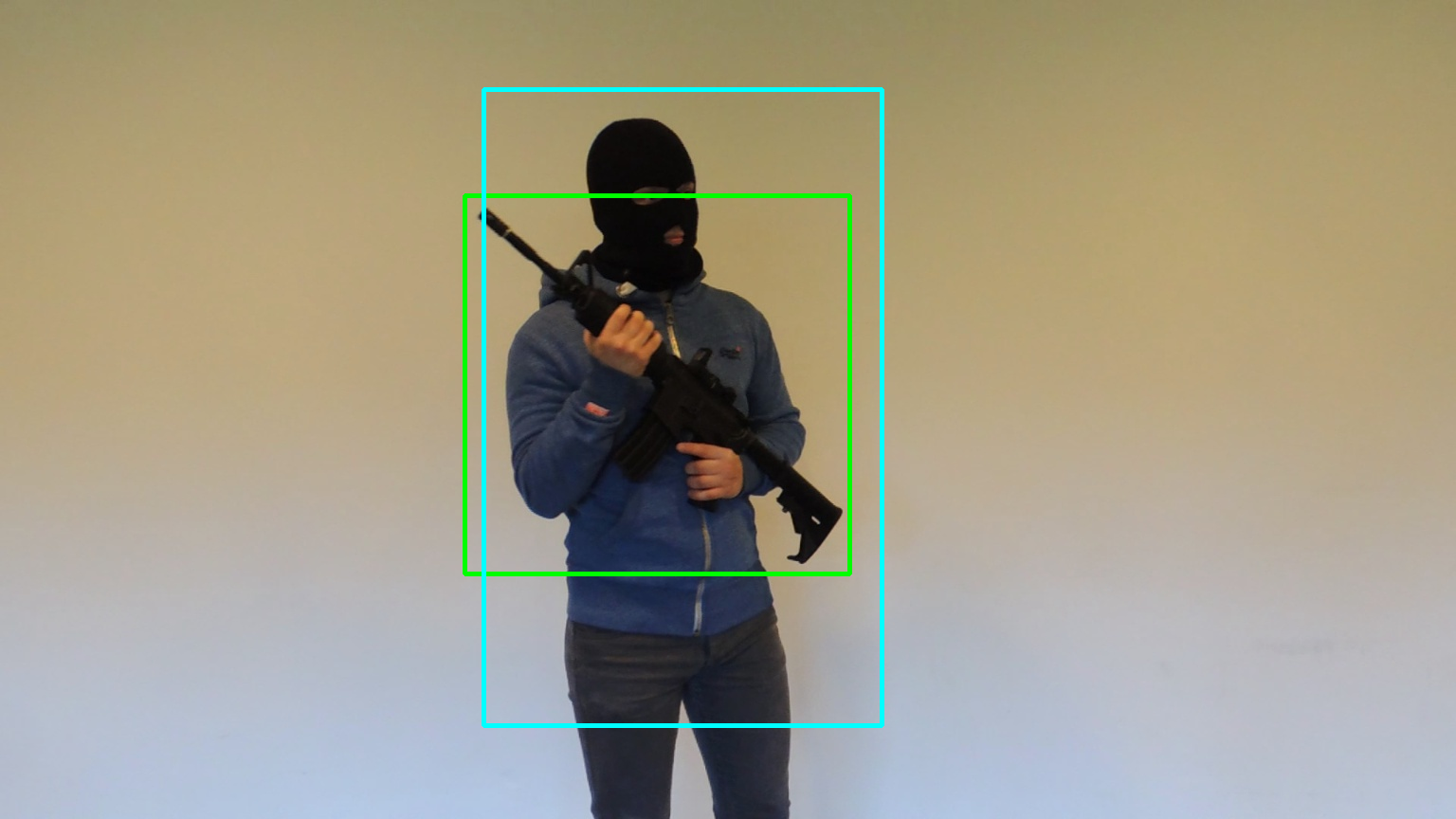}
\end{figure}

\begin{figure}[h]
\caption{Result of the simulation testing with rifle being held in ``low" position. The green bounding box is the manually labeled area, whereas the blue bounding box is the area detected by the networks. \label{fig:video2}}
\includegraphics[width=\linewidth]{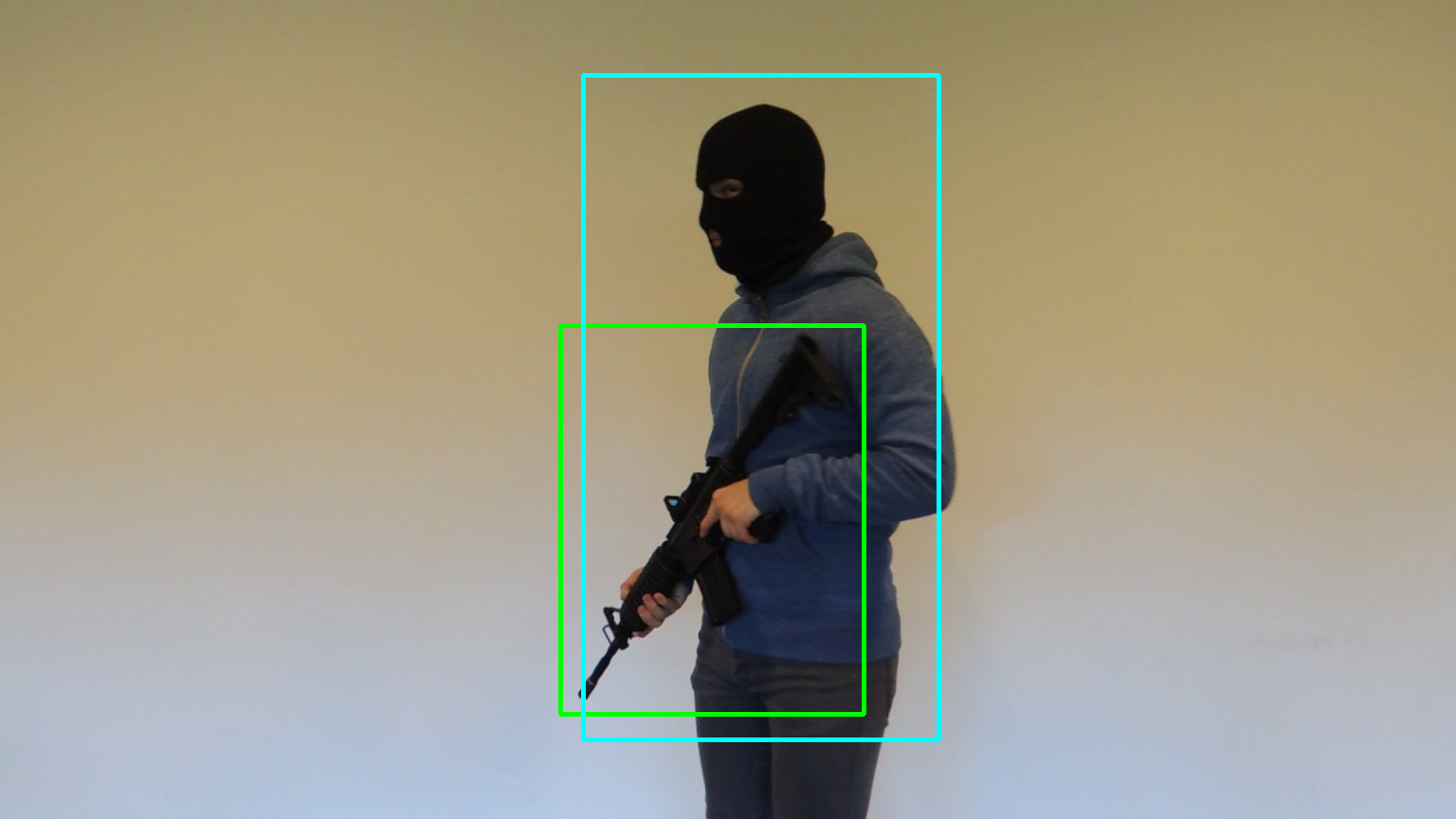}
\end{figure}

Figure \ref{fig:video1} shows a a successful detection with an accurate computed bounding box. Figure \ref{fig:video2} shows the case of a successful detection, but where the bounding box was mistakenly computed and included the head of the person holding the weapon.
 
%(Numbers and graphs)
\paragraph{Discussion}
%(Interpretation, meaning and consequences of the results)
In this simulation, the system had the task of identifying a AR-15 rifle held by a person. Besides different orientations and angles, this scenario takes into consideration concrete distractors, such as the person itself and its clothes. The results showed that semantic segmentation modelling is indeed promising in detecting a firearm in different angles and positions, but nevertheless, it is still prone to misclassifying objects that bear similar color or shape to the ones that have been used to train the model. For instance, the misclassification of the black balaclava as shown on Figure \ref{fig:video2} suggests that the negative set used during the training should include a wide variety of images to cover all possible situations and environmental features such as clothing, headwear and background details.

\section{Ethical considerations}\label{sec:EthicalConsiderations}
Use of machine learning in safety-critical situations raises ethical questions about the uses and potential misuses of learning algorithms. In our case, while our aim was to develop an effective system that may be used to help security personnel in detecting threatening situations in civil society and in intervening more promptly to prevent loss of lives, we are aware that our system may be tuned and deployed in other scenarios; in particular, it is not far-fetched to see its use within lethal autonomous weapon system (LAWS). The authors, however, discourage and condemn the use of their system in autonomous weapons\footnote{\url{https://futureoflife.org/open-letter-autonomous-weapons/}}.

To prevent our system from making decision autonomously, we also considered a potential extension, including a human-in-the-loop. In the context of critical AI systems designed to make potentially dangerous decisions, a simple solution adopted to prevent catastrophic errors is offered by integrating a human overseer in the decision loop of the machine \cite{saunders2017trial,abel2017agent}.
We considered the implementation of a human-in-the-loop system both at training time and at test time. During training, such an approach would allow a human supervisor to check and correct the actual decision of the system, it would allow the detection of common mistakes and biases in the network, and it would make it possible to transfer human insights by providing a stronger learning signal to the machine. After deployment, a human overseer would check and validate the outputs of the network and would be responsible and accountable for the final decision. 

%To implement this technique, we allowed a human controller to check the detection output of the network and intervene in case of false classification. Operator had an ability to select the correct area containing a firearm part in question and parse it as a positive example for training, while false classified sample was used as the negative one. This not only allowed for the ongoing correction of training, but also gave an opportunity to dynamically enrich datasets.

\section{Conclusions and future work}\label{sec:Conclusions}
%(Overall conclusion of this work. Potential directions for future work: you can refer work in your thesis that did not find a place here, or write about other improvements like tuning the network architecture, generating more realistic data, integrating in a more complex semantic model, training using zooming...)
%(Around .5 pages)
In this paper, we considered the problem of detecting the presence of weapons within images or video frames. We suggested a solution based on machine learning techniques and the principle of semantic decomposition of the problem. Instead of implementing a single complex learning module which would require large amounts of data and high computational resources, we proposed the use of multiple simple neural networks that could be cheaply trained to detect only specific components of a weapon and which could be easily aggregated to produce robust output. We tested our model on data from the same distribution as the training data (i.e., Google images), we examined it on synthetic sanitized data, and we evaluated it on out-of-sample (i.e., video frames) data. The results showed the reliability of the individual component part networks and the versatility of the overall system in integrating the outputs of the single networks.

While promising, our results are still preliminary and different directions of future developments are left open. From a practical point of view, the system may be further refined by generating more extensive and representative data sets and by considering additional components of a rifle. Another obvious extension would be to enlarge the set of detected objects from the simple AR-15  model to other rifles or, even, other weapons, such as knives.
The process of aggregating the outputs of individual networks could be refined as well. In this work, we proposed a simple majority rule, and we suggested in passing other potential rules. However, the problem of producing a single output may be more formally expressed in probabilistic terms, and we could exploit other statistical models to generate outputs reporting not only a final decision but their uncertainty, as well. This work could better reconnect our research to the theory of ensemble learning as well.

Most importantly, we are considering the development of a thorough comparison between our semantic segmentation approach and standard approaches presented in the literature and based on a single deeper neural network able to identify at once the presence of weapons in an image. A fair comparison is far from trivial, as it would require the training of models with similar capacity and using equally informative data sets. We are designing such experiments with the aim to precisely evaluate the strengths and the weaknesses of the two approaches in terms of accuracy, confusion matrices, and training cost in terms of number of samples and computational time.

\section*{Acknowledgment}
This research was supported by the research project Oslo Analytics funded by the Research Council of Norway under the Grant No.: IKTPLUSS 247648.
\balance
\bibliographystyle{ieeetr}
\bibliography{references}

\begin{thebibliography}{10}

\bibitem{UNODC2019}
{United Nations Office on Drugs and Crime (UNODC)}, ``Global study on homicide
  2019. data: Unodc homicide statistics 2019.''
  \url{https://www.unodc.org/unodc/en/data-and-analysis/global-study-on-homicide.html},
  July 2019.

\bibitem{Darker2017_CCTV}
I.~{Darker}, A.~{Gale}, L.~{Ward}, and A.~{Blechko}, ``Can cctv reliably detect
  gun crime?,'' in {\em 2007 41st Annual IEEE International Carnahan Conference
  on Security Technology}, pp.~264--271, Oct 2007.

\bibitem{bishop2006pattern}
C.~M. Bishop, {\em Pattern recognition and machine learning}.
\newblock springer, 2006.

\bibitem{lecun2015deep}
Y.~LeCun, Y.~Bengio, and G.~Hinton, ``Deep learning,'' {\em nature}, vol.~521,
  no.~7553, p.~436, 2015.

\bibitem{krizhevsky2012imagenet}
A.~Krizhevsky, I.~Sutskever, and G.~E. Hinton, ``Imagenet classification with
  deep convolutional neural networks,'' in {\em Advances in neural information
  processing systems}, pp.~1097--1105, 2012.

\bibitem{GUO201627}
Y.~Guo, Y.~Liu, A.~Oerlemans, S.~Lao, S.~Wu, and M.~S. Lew, ``Deep learning for
  visual understanding: A review,'' {\em Neurocomputing}, vol.~187, pp.~27 --
  48, 2016.
\newblock Recent Developments on Deep Big Vision.

\bibitem{Krizhevsky2012}
A.~Krizhevsky, I.~Sutskever, and G.~E. Hinton, ``Imagenet classification with
  deep convolutional neural networks,'' in {\em Proceedings of the 25th
  International Conference on Neural Information Processing Systems - Volume
  1}, NIPS'12, (USA), pp.~1097--1105, Curran Associates Inc., 2012.

\bibitem{Xue2002}
Z.~{Xue}, R.~S. {Blum}, and Y.~{Li}, ``Fusion of visual and ir images for
  concealed weapon detection,'' in {\em Proceedings of the Fifth International
  Conference on Information Fusion. FUSION 2002. (IEEE Cat.No.02EX5997)},
  vol.~2, pp.~1198--1205 vol.2, July 2002.

\bibitem{Sheen2001}
D.~M. {Sheen}, D.~L. {McMakin}, and T.~E. {Hall}, ``Three-dimensional
  millimeter-wave imaging for concealed weapon detection,'' {\em IEEE
  Transactions on Microwave Theory and Techniques}, vol.~49, pp.~1581--1592,
  Sep. 2001.

\bibitem{Dalal2005_top}
N.~{Dalal} and B.~{Triggs}, ``Histograms of oriented gradients for human
  detection,'' in {\em 2005 IEEE Computer Society Conference on Computer Vision
  and Pattern Recognition (CVPR'05)}, vol.~1, pp.~886--893 vol. 1, June 2005.

\bibitem{Felzenszwalb2010}
P.~F. {Felzenszwalb}, R.~B. {Girshick}, D.~{McAllester}, and D.~{Ramanan},
  ``Object detection with discriminatively trained part-based models,'' {\em
  IEEE Transactions on Pattern Analysis and Machine Intelligence}, vol.~32,
  pp.~1627--1645, Sep. 2010.

\bibitem{felzenszwalb2005pictorial}
P.~F. Felzenszwalb and D.~P. Huttenlocher, ``Pictorial structures for object
  recognition,'' {\em International journal of computer vision}, vol.~61,
  no.~1, pp.~55--79, 2005.

\bibitem{Hosang2016}
J.~Hosang, R.~Benenson, P.~Dollar, and B.~Schiele, ``What makes for effective
  detection proposals?,'' {\em IEEE Trans. Pattern Anal. Mach. Intell.},
  vol.~38, pp.~814--830, Apr. 2016.

\bibitem{Girshick2014}
R.~{Girshick}, J.~{Donahue}, T.~{Darrell}, and J.~{Malik}, ``Rich feature
  hierarchies for accurate object detection and semantic segmentation,'' in
  {\em 2014 IEEE Conference on Computer Vision and Pattern Recognition},
  pp.~580--587, June 2014.

\bibitem{uijlings2013selective}
J.~R. Uijlings, K.~E. Van De~Sande, T.~Gevers, and A.~W. Smeulders, ``Selective
  search for object recognition,'' {\em International journal of computer
  vision}, vol.~104, no.~2, pp.~154--171, 2013.

\bibitem{NIPS2015_5638}
S.~Ren, K.~He, R.~Girshick, and J.~Sun, ``Faster r-cnn: Towards real-time
  object detection with region proposal networks,'' in {\em Advances in Neural
  Information Processing Systems 28}, pp.~91--99, Curran Associates, Inc.,
  2015.

\bibitem{gelana2019firearm}
F.~Gelana and A.~Yadav, ``Firearm detection from surveillance cameras using
  image processing and machine learning techniques,'' in {\em Smart Innovations
  in Communication and Computational Sciences}, pp.~25--34, Springer, 2019.

\bibitem{olmos2018automatic}
R.~Olmos, S.~Tabik, and F.~Herrera, ``Automatic handgun detection alarm in
  videos using deep learning,'' {\em Neurocomputing}, vol.~275, pp.~66--72,
  2018.

\bibitem{rokach2010pattern}
L.~Rokach, {\em Pattern classification using ensemble methods}, vol.~75.
\newblock World Scientific, 2010.

\bibitem{NEWZEALAND2018}
{National Public Radio, Inc. (NPR)}, ``New zealand starts gun buyback program
  in response to christchurch mosque shootings.''
  \url{https://www.npr.org/2019/07/13/741480261/}, July 2019.
\newblock Accessed: 28 July 2019.

\bibitem{LASVEGAS2017}
``Setting sights on the ar-15: After las vegas shooting, lawyers target gun
  companies.''

\bibitem{ORLANDO2016}
{NBC News}, ``Ar-15 style rifle used in orlando massacre has bloody pedigree.''
  \url{https://www.nbcnews.com/storyline/orlando-nightclub-massacre}, June
  2016.
\newblock Accessed: 01 August 2019.

\bibitem{Srivastava2013a}
N.~Srivastava, {\em Improving neural networks with dropout}.
\newblock PhD thesis, University of Toronto, 2013.

\bibitem{saunders2017trial}
W.~Saunders, G.~Sastry, A.~Stuhlm\"{u}ller, and O.~Evans, ``Trial without
  error: Towards safe reinforcement learning via human intervention,'' in {\em
  Proceedings of the 17th International Conference on Autonomous Agents and
  MultiAgent Systems}, AAMAS '18, (Richland, SC), pp.~2067--2069, International
  Foundation for Autonomous Agents and Multiagent Systems, 2018.

\bibitem{abel2017agent}
D.~Abel, J.~Salvatier, A.~Stuhlmüller, and O.~Evans, ``Agent-agnostic
  human-in-the-loop reinforcement learning,'' in {\em NIPS 2016 Workshop:
  Future of Interactive Learning Machines}, Barcelona, Spain 2016.

\end{thebibliography}

\end{document}